\documentclass[runningheads]{llncs}
\usepackage{graphicx}
\usepackage{booktabs}
\usepackage{multirow}
\usepackage{algorithm}
\usepackage[noend]{algpseudocode}

\algrenewcommand{\algorithmiccomment}[1]{\hskip3px$\#$ #1}
\algnewcommand\algorithmicforeach{\textbf{for each}}
\algdef{S}[FOR]{ForEach}[1]{\algorithmicforeach\ #1\ \algorithmicdo}
\algdef{SE}[SUBALG]{Indent}{EndIndent}{}{\algorithmicend\ }
\algtext*{Indent}
\algtext*{EndIndent}

\usepackage{orcidlink}

\hypersetup{
    colorlinks=true,
    linkcolor=purple,
    filecolor=cyan,      
    urlcolor=teal,
    citecolor=blue,
    }

\begin{document}

\title{Classy Ensemble:
       \texorpdfstring{\\A Novel Ensemble Algorithm for Classification}{}}

% \author{Moshe Sipper\inst{1}\orcidID{0000-0003-1811-472X}}
\author{Moshe Sipper\inst{1}\orcidlink{0000-0003-1811-472X}}
\authorrunning{M. Sipper}
\institute{Department of Computer Science, Ben-Gurion University, Beer-Sheva 84105, Israel
\email{sipper@bgu.ac.il}, 
\url{http://www.moshesipper.com/}}

\maketitle

\begin{abstract}
We present \textit{Classy Ensemble}, a novel ensemble-generation algorithm for classification tasks, which aggregates models through a weighted combination of per-class accuracy. Tested over 153 machine learning datasets we demonstrate that Classy Ensemble outperforms two other well-known aggregation algorithms---order-based pruning and clustering-based pruning---as well as the recently introduced lexigarden ensemble generator. 
We then present three enhancements: 
1) Classy Cluster Ensemble, which combines Classy Ensemble and cluster-based pruning;
2) Deep Learning experiments, showing the merits of Classy Ensemble over four image datasets: Fashion MNIST, CIFAR10, CIFAR100, and ImageNet;
and 
3) Classy Evolutionary Ensemble, wherein an evolutionary algorithm is used to select the set of models which Classy Ensemble picks from. 
This latter, combining learning and evolution, resulted in improved performance on the
hardest dataset.

\keywords{Ensemble learning  \and Classification \and Machine learning \and Deep learning \and ImageNet.}
\end{abstract}

\section{Introduction}
\label{sec:intro}
Ensemble techniques---which aggregate the outputs of several models to produce a final prediction---are prevalent nowadays within the field of machine learning (ML) and deep learning (DL). Well-known methods such as bagging \cite{breiman1996bagging}, boosting \cite{freund1995desicion}, and stacking \cite{wolpert1992stacked} have been around for decades.
These traditionally rely on the aggregation of many weak models.

In more recent years the merits of combining strong models have been shown, which serves as the starting point for our work herein. We will present a novel ensemble-generation algorithm for classification tasks, called \textit{Classy Ensemble}, which aggregates  models by examining their per-class accuracy. Through extensive experimentation we will show that our new algorithm is often able to improve upon top ML algorithms. We shall also show strong results for DL models.

The next section discusses ensemble learning. In Section~\ref{sec:classy} we describe our novel algorithm: Classy Ensemble . We delineate the experimental setup in Section~\ref{sec:setup}, followed by results in Section~\ref{sec:results}, and a discussion in Section~\ref{sec:discussion}. Further experimentation on three enhancements is presented in Section~\ref{sec:enhancements}, ending with concluding remarks in Section~\ref{sec:conc}.

% Note that this paper focuses on ML models and datasets---in the future we plan to expand to DL as well (Section~\ref{sec:conc}).

\section{Ensemble Learning}
\label{sec:ensemble}
The field of ensemble learning has an illustrious history spanning several decades, and it is beyond the scope of this paper to survey it all. In several recent works we presented novel ensemble algorithms and we refer the reader to literature surveys therein (in particular, our most recent work \cite{Sipper2023Combining}):
\begin{itemize}
    \item In \cite{Sipper2020,Sipper2021cml} we presented \textit{conservation machine learning}, which conserves models across runs, users, and experiments. As part of this work we compared multiple ensemble-generation methods, also introducing \textit{lexigarden}, discussed below.

    \item \cite{sipper2021symbolic} presented SyRBo---Symbolic-Regression Boosting---an ensemble method based on strong learners that are combined via boosting, used to solve regression tasks.

    \item \cite{Sipper2021addgBoost} introduced AddGBoost, a gradient boosting-style algorithm, wherein the decision tree is replaced by a succession of stronger learners, which are optimized via a state-of-the-art hyperparameter optimizer.

    \item \cite{Sipper2023Combining} presented a comprehensive, stacking-based framework for combining deep learning with good old-fashioned machine learning, called Deep GOld. The framework involves ensemble selection from 51 retrained pretrained deep networks as first-level models, and 10 ML algorithms as second-level models.
\end{itemize}

We continue in the vein of the above works, maintaining ensemble component models that are not weak, as usually done with bagging and boosting methods (going back to the seminal paper of Schapire, ``The strength of weak learnability'' \cite{schapire1990strength}). On the contrary, we aim to find strong models, to be improved upon through ``intelligent'' aggregation.

Ultimately, we chose to implement three ensemble-generation methods, which will be compared with our new Classy Ensemble (all code described herein is available at \url{https://github.com/moshesipper/classy-ensemble}):
\begin{enumerate}
    \item \textit{Order-based pruning.} We implemented two methods of ensemble pruning \cite{Sipper2021cml,tsoumakas2009ensemble}. The first, order-based, sorts all available models from best model to worst (according to validation score), and then selects the top $k$ models for the ensemble, with $k$ being a specified hyperparameter. We denote this \textit{Order Ensemble}.
    
    \item \textit{Clustering-based pruning.} The second ensemble-pruning method performs $k$-means clustering over all model validation output vectors, with a given number of clusters, $k$, and then produces an ensemble by collecting the top-scoring model of each cluster. We denote this \textit{Cluster Ensemble} (for more details see \cite{Sipper2021cml}). 
    
    The above two ensemble algorithms were implemented as part of our research in \cite{Sipper2021cml}, and re-implemented herein, with some programmatic improvements.
    
    \item \textit{Lexigarden}, introduced in \cite{Sipper2021cml}, receives a collection of models, a dataset along with target values, and the number of models the generated ``garden'' (ensemble) is to contain---\textit{garden size}. Lexigarden performs the following operations \textit{garden size} times, to obtain en ensemble of models:
    randomly shuffle the dataset, then successively iterate through it, retaining only the models that provide a correct answer for each instance; ultimately we are left with either a single model or a small number of models (from which one is chosen at random)---these are precisely those models that have correctly classified the subset of the dataset engendered through the looping process. 
    
    Lexigarden thus selects a diverse subset of all the models by picking ones that each excels on a particular random subset of instances. More details about lexigarden can be found in \cite{Sipper2021cml}.

\end{enumerate}

\section{Classy Ensemble}
\label{sec:classy}

Algorithm~\ref{alg:basic} implements basic ensemble functionality. As such, Order-based pruning (Order Ensemble) is obtained through the \textit{topk} parameter of the \textsc{init} method. The class \texttt{BasicEnsemble} is also used by classes \texttt{ClusterEnsemble} and \texttt{Lexigarden}, which implement, respectively, Cluster Ensemble and Lexigarden.

\begin{algorithm}[hbt!]
\small
\caption{BasicEnsemble}\label{alg:basic}
\begin{algorithmic}[1]

\Function{init}{\textit{models}, \textit{topk=None}}
\Statex\Comment{\textit{models}: list of fitted models}
\Statex\Comment{\textit{topk}: how many top models to include in ensemble}
    \If{\textit{topk is None}}
       \State \textit{ensemble} $\gets$ all \textit{models}
    \Else
       \State \textit{ensemble} $\gets$  \textit{topk} \textit{models}
    \EndIf
\EndFunction

\Statex

\Function{predict}{\textit{X}}
    \State \textit{predictions} $\gets$ concatenate predictions of all models in \textit{ensemble} over data \textit{X}
    \State Compute and return majority prediction of each row (sample) in \textit{predictions}
\EndFunction
\end{algorithmic}
\normalsize
\end{algorithm}

Algorithm~\ref{alg:classy} delineates our new ensemble algorithm, Classy Ensemble. 
It receives as input a collection of fitted models, each one's overall accuracy score, and per-class accuracy, i.e., each model's accuracy values as computed separately for every class (note: we used \texttt{scikit-learn}'s \cite{scikit-learn} \texttt{balanced\_accuracy\_score}, which avoids inflated performance estimates on imbalanced datasets).

\begin{algorithm}[hbt!]
\small
\caption{ClassyEnsemble}\label{alg:classy}
\begin{algorithmic}[1]

\Function{init}{\textit{models}, \textit{n\_classes}, \textit{topk}}
\Statex\Comment{\textit{models}: list of fitted models}
\Statex\Comment{\textit{n\_classes}: number of classes in dataset}
\Statex\Comment{\textit{topk}: how many top models to include in ensemble}

\State \textit{ensemble} = \{\}
\For{\textit{c} $\gets$ 1 to \textit{n\_classes}} \label{line:for-c}
\State \textit{sorted} $\gets$ sort \textit{models} according to validation accuracy over class $c$ \label{line:sort-c}
  \ForAll {\textit{model} $\in$ \textit{topk} of \textit{sorted}}
     \If {\textit{model} $\notin$ \textit{ensemble}}
        \State Add \textit{model} to \textit{ensemble}
     \EndIf
     \State Mark \textit{model} as voter for class $c$
  \EndFor
\EndFor
    
\EndFunction

\Statex

\Function{predict}{\textit{X}}
  \State \textit{predictions} $\gets$ $0$ \Comment{vector of appropriate dimensions}
  \ForAll {\textit{model} $\in$ \textit{ensemble}}
     \State \textit{sc} $\gets$ \textit{model} overall validation score
     \State \textit{cls} $\gets$ \textit{model} voter vector
     \State \textit{p} $\gets$ \textit{model}.predict\_proba(X) \Comment{prediction probabilities per class}
     \State \textit{predictions} +=  \textit{p} * \textit{sc} * \textit{cls}  \Comment{\textit{p} is a matrix of dimensions \textit{number of samples} $\times$ \textit{n\_classes}, \textit{sc} is a scalar, \textit{cls} is a vector of size \textit{n\_classes}; the matrix multiplication uses Python broadcasting}
  \EndFor
  \State Return majority votes (per each sample) of \textit{predictions}
\EndFunction
\end{algorithmic}
\normalsize
\end{algorithm} 

As seen in Algorithm~\ref{alg:classy}, Classy Ensemble adds to the ensemble the \textit{topk} best-performing (over validation set) models, \textit{for each class}. A model may be in the \textit{topk} set for more than one class. Thus, for each model in the ensemble, we also maintain a list of classes for which it is a \textit{voter}, i.e., its output for each voter class is taken into account in the final aggregation. 
The binary voter vector of size \textit{n\_classes} is set to 1 for classes the model is permitted to vote for, 0 otherwise.

Thus, a model not in the ensemble is obviously not part of the final aggregrated prediction; further, a model \textit{in} the ensemble is only ``allowed'' to vote for those classes for which it is a voter---i.e., for classes it was amongst the \textit{topk}.

Classy Ensemble provides a prediction by aggregating its members' predicted-class probabilities, weighted by the overall validation score, and taking into account voter permissions.

\section{Experimental Setup}
\label{sec:setup}

To test our new algorithm and compare it against the three benchmark ensemble algorithms described above (Order, Cluster, Lexigarden), we used the recently introduced PMLB repository of datasets \cite{romano2021pmlb}. A full experiment over a single dataset consisted of 30 replicate runs (described below). Of the 162 classification datasets in PMLB, we were able to complete full experiments for 153 datasets (we imposed a 10-hour limit to avoid overly lengthy runs, so that we could collect a sizeable chunk of results to analyze). The number of samples was in the range 32 -- 58,000, the number of features was in the range 2 -- 1000, and the number of classes was in the range 2 -- 18.

Algorithm~\ref{alg:setup} details the experimental setup per dataset.
Each replicate fits 250 randomly chosen models (from a set of 8 algorithms) to the dataset. For each model, the dataset is split randomly into training, validation, and test sets, the model is trained on the training set, and then validated on the validation set. The model with the highest validation score is tested on the test set. 

\begin{algorithm}[hbt!]
\small
\caption{Experimental setup (per dataset)}\label{alg:setup}
\begin{algorithmic}[1]
\Statex
\Require
\Indent
\Statex \textit{dataset} $\gets$ dataset to be used
\Statex \textit{n\_replicates} $\gets$ 30
\Statex \textit{n\_models} $\gets$ 250
\Statex \textit{topk} $\gets$ [1, 2, 3, 5, 20, 50, 100, n\_models]
\Statex 
\textit{models} $\gets$ \{\texttt{LogisticRegression}, \texttt{SGDClassifier}, \texttt{DecisionTreeClassifier}, \texttt{RandomForestClassifier}, \texttt{AdaBoostClassifier}, \texttt{KNeighborsClassifier}, \texttt{XGBClassifier}, \texttt{LGBMClassifier}\}

\EndIndent

\Ensure
\Indent
\Statex Statistics (over all replicates)
\EndIndent

\Statex
\For{\textit{rep} $\gets$ 1 to \textit{n\_replicates}} 
  \For{\textit{model$_i$} $\gets$ 1 to \textit{n\_models}}  
     \State Randomly split \textit{dataset} into \textit{training}, \textit{validation}, and \textit{test} sets \Comment{60-20-20\%}     
     \State Fit \texttt{StandardScaler} to \textit{training} set and apply fitted scaler to \textit{validation} and \textit{test} sets
     \State Select \textit{model} at random from \textit{models}
     \State Train \textit{model} over training set and compute score over \textit{validation} set
  \EndFor
  
  \State Test \textit{model} with best validation score over \textit{test} set

  \For{\textit{ensemble} in [OrderEnsemble, ClassyEnsemble, Lexigarden, ClusterEnsemble]}  
    \For{\textit{k} in \textit{topk}} \label{line:beginfor}
       \State\parbox[t]{.8\textwidth}{ 
       Generate \textit{ensemble} with parameter \textit{k} \Comment{\textit{k} is top models for BasicEnsemble and ClassyEnsemble, garden size for Lexigarden, and number of clusters for ClusterEnsemble; validation set is used in ensemble generation}
       }
    \EndFor
    \State \textit{best ensemble} $\gets$ \textit{ensemble} with highest validation score over \textit{k} $\in$ \textit{topk}  \label{line:bestk}
    \State Test \textit{best ensemble} over \textit{test} set
  \EndFor  
\EndFor

\State Compute test-set statistics over all replicates
\end{algorithmic}
\normalsize
\end{algorithm}

The experiment then proceeds to generate ensembles and assess their performance. For each of the four ensemble methods, an ensemble is generated with hyperparameter $k$ (top models, garden size, or number of clusters---depending on the ensemble method), with 8 values of $k$ being used: $[1, 2, 3, 5, 20, 50, 100, 250]$. The ensemble that performs best on the validation data is tested on the test data. 

Note that we divided the training data further into training and validation sets, using the former to train the models, and the latter to pick the best model, as well as to compose the ensembles. This provides a better assessment of generalization capabilities, while still using what is essentially training data (i.e., no data leakage).
The test data is used only at the very end for final model or ensemble evaluation.

\section{Results}
\label{sec:results}
Table~\ref{tab:wins} shows the number of statistically significant ``wins'', i.e., datasets whose 30-replicate experiment produced ensembles that bested the top single-model performer. We assessed statistical significance by performing a 10,000-round permutation test, comparing the median (test-set) scores of the replicates' single top models vs. ensembles; if the ensemble's median score was higher and the p-value was $< 0.05$, the improvement was considered significant.

\begin{table}[hbt!]
\begin{center}
\caption{Summary of results over 153 dataset experiments, as described in Section~\ref{sec:setup} and delineated in Algorithm~\ref{alg:setup}. 
The statistics reported are for the test-set scores. 
Wins: Number of datasets for which the ensemble method bested the best single model in a statistically significant manner.
Unique: Number of Wins for which the ensemble method was the only one to improve upon best single model.
Size: Mean (SD) of median ensemble sizes per experiment.
Best $k$: Mean (SD) of median best-$k$ values per experiment, i.e., $k$ values that produced best ensembles (Algorithm~\ref{alg:setup}, lines \ref{line:beginfor}--\ref{line:bestk})
}\label{tab:wins}%
\begin{tabular}{@{}r@{\hskip 5pt}c@{\hskip 5pt}c@{\hskip 5pt}c@{\hskip 10pt}c@{}}
\toprule
Ensemble & Wins  & Unique & Size & Best $k$\\
\midrule
Classy Ensemble  & 45 & 25 & 56 (69) & 27 (45) \\
Lexigarden       & 22  & 9 & 32 (32) & 32 (32) \\
Order Ensemble   & 12 & 2  & 40 (44) & 40 (44) \\
Cluster Ensemble & 10 & 1  & 2.7 (1) & 2.7 (1) \\
\bottomrule
\end{tabular}
\end{center}
\end{table}

Note that in Table~\ref{tab:wins} `Size' and `Best $k$' are equal for all but Classy Ensemble. This is to be expected since ensemble size matches the $k$ value, except for Classy Ensemble, wherein $k$ specifies selection of top $k$ models \textit{per class}, and thus the total number of models is usually larger.

Table~\ref{tab:bestmodels} shows the 45 datasets for which Classy Ensemble's win was significant, along with the models it won against.

\begin{table}[hbt!]
\scriptsize
\begin{center}
\caption{Left: Datasets for which Classy Ensemble improved upon the best single models in a statistically significant manner.
Right: Best single models with win counts, over the 30 replicate runs}\label{tab:bestmodels}%
\begin{tabular}{@{}r@{\hskip 7pt}l@{}}
\toprule
Dataset & Best Single Models \\
\midrule
allbp & {DecisionTreeClassifier: 30} \\
allhyper & {XGBClassifier: 30} \\
analcatdata\_boxing2 & {SGDClassifier: 30} \\
analcatdata\_germangss & {SGDClassifier: 22, LGBMClassifier: 8} \\
analcatdata\_lawsuit & {RandomForestClassifier: 25, SGDClassifier: 5} \\
ann\_thyroid & {LGBMClassifier: 27, RandomForestClassifier: 3} \\
auto & {XGBClassifier: 30} \\
cleveland\_nominal & {RandomForestClassifier: 29, SGDClassifier: 1} \\
cmc & {SGDClassifier: 30} \\
colic & {SGDClassifier: 26, RandomForestClassifier: 3, KNeighborsClassifier: 1} \\
dis & {DecisionTreeClassifier: 30} \\
dna & {XGBClassifier: 30} \\
ecoli & {KNeighborsClassifier: 29, SGDClassifier: 1} \\
flags & {SGDClassifier: 29, LogisticRegression: 1} \\
Epistasis...1000... & {KNeighborsClassifier: 27, RandomForestClassifier: 3} \\
Epistasis...20... & {LGBMClassifier: 28, RandomForestClassifier: 2} \\
Heterogeneity...20... & {LGBMClassifier: 25, RandomForestClassifier: 5} \\
heart\_h & {SGDClassifier: 26, DecisionTreeClassifier: 3, RandomForestClassifier: 1} \\
heart\_statlog & {SGDClassifier: 29, RandomForestClassifier: 1} \\
Hill\_Valley... & {SGDClassifier: 25, DecisionTreeClassifier: 5} \\
hypothyroid & {LGBMClassifier: 30} \\
ionosphere & {DecisionTreeClassifier: 30} \\
iris & {SGDClassifier: 29, RandomForestClassifier: 1} \\
kr\_vs\_kp & {RandomForestClassifier: 16, DecisionTreeClassifier: 14} \\
led7 & {SGDClassifier: 30} \\
lymphography & {SGDClassifier: 29, AdaBoostClassifier: 1} \\
magic & {RandomForestClassifier: 29, XGBClassifier: 1} \\
mfeat\_fourier & {RandomForestClassifier: 30} \\
mfeat\_morphological & {LogisticRegression: 25, SGDClassifier: 5} \\
mfeat\_pixel & {KNeighborsClassifier: 25, RandomForestClassifier: 5} \\
molecular\_biology... & {RandomForestClassifier: 19, AdaBoostClassifier: 11} \\
pendigits & {RandomForestClassifier: 25, LGBMClassifier: 5} \\
phoneme & {RandomForestClassifier: 30} \\
prnn\_fglass & {RandomForestClassifier: 17, DecisionTreeClassifier: 12, SGDClassifier: 1} \\
prnn\_synth & {DecisionTreeClassifier: 27, SGDClassifier: 3} \\
satimage & {LGBMClassifier: 30} \\
solar\_flare\_1 & {SGDClassifier: 27, LGBMClassifier: 3} \\
solar\_flare\_2 & {RandomForestClassifier: 26, SGDClassifier: 4} \\
splice & {LGBMClassifier: 30} \\
twonorm & {SGDClassifier: 28, RandomForestClassifier: 2} \\
waveform\_21 & {LogisticRegression: 19, SGDClassifier: 11} \\
waveform\_40 & {SGDClassifier: 28, RandomForestClassifier: 2} \\
wdbc & {KNeighborsClassifier: 21, RandomForestClassifier: 7, SGDClassifier: 2} \\
wine\_quality\_red & {LGBMClassifier: 16, RandomForestClassifier: 13, SGDClassifier: 1} \\
wine\_quality\_white & {RandomForestClassifier: 30} \\
\bottomrule
\end{tabular}
\end{center}
\normalsize
\end{table}

Table~\ref{tab:ds} shows dataset properties for the wins of each method.

\begin{table}[hbt!]
\begin{center}
\caption{Dataset properties for the significant wins of each method (Table~\ref{tab:wins}).
Samples: Median number of samples.
Features: Median number of features.
Classes: Median number of classes.
}\label{tab:ds}%
\begin{tabular}{@{}r@{\hskip 7pt}ccc@{}}
\toprule
Ensemble & Samples  & Features & Classes \\
\midrule
Classy Ensemble  & 1600 & 20   & 3 \\
Lexigarden       & 690  & 22   & 2  \\
Order Ensemble   & 1473 & 16.5 & 2.5  \\
Cluster Ensemble & 951  & 27   & 2  \\
\bottomrule
\end{tabular}
\end{center}
\end{table}

\section{Discussion}
\label{sec:discussion}

Table~\ref{tab:wins} demonstrates the superiority of Classy Ensemble over other ensemble-generation methods, outperforming them by a wide margin. Further, it is able to improve results in a statistically significant manner on close to one third of the datasets. 

Examining Table~\ref{tab:bestmodels}, we observe that Classy Ensemble often beats top ML algorithms, such as XGBoost, LightGBM, Random Forest, and SGD. This adds strength to our argument regarding the potency of the newly proposed method.

Table~\ref{tab:ds} attempts to discern whether Classy Ensemble is better on some types of datasets than others. Conclusions from this table should be taken with a grain of salt as the number of wins is particularly small for Order Ensemble and Cluster Ensemble. That said, Classy Ensemble seems to be able to handle larger datasets with more classes.

\section{Enhancements}
\label{sec:enhancements}
\textbf{Classy Cluster Ensemble.} We next combined two of the above methods---the newly introduced Classy Ensemble and cluster-based pruning---to form Classy Cluster Ensemble:
The model outputs are clustered, and the per-class sorting of Classy Ensemble is done per cluster, with the top models per class, per cluster, added to the ensemble. Technically, an additional outer loop is added above line~\ref{line:for-c} in Algorithm~\ref{alg:classy}, looping over all clusters, and the sorting in line~\ref{line:sort-c} is done over the cluster models rather than all models. While this proved effective, it still came in second to Classy Ensemble, with 38 significant datasets wins. We tested a second version of Classy Cluster Ensemble, wherein instead of maintaining a single global ensemble, we maintain an ensemble per cluster, but results were poor. 

\textbf{Deep Learning.} We experimented with deep networks by adapting the framework to run with PyTorch \cite{paszke2019pytorch}, using 51 models we had trained in \cite{Sipper2023Combining} (see Section~\ref{sec:ensemble}).
Four image datasets were employed (Table~\ref{tab:dl-datasets}): 
Fashion MNIST, CIFAR10, CIFAR100, and ImageNet. Note that for these  datasets there are public, given training and test sets.

\begin{table}[hbt!]
\begin{center}
\caption{DL datasets
}\label{tab:dl-datasets}%
\begin{tabular}{@{}r@{\hskip 7pt}cccc@{}}
\toprule
Dataset & Images & Classes & Training & Test \\ 
\midrule
Fashion MNIST & 28x28 grayscale & 10 & 60,000 & 10,000 \\
CIFAR10 & 32x32 color & 10 & 50,000 & 10,000 \\
CIFAR100 & 32x32 color & 100 & 50,000 & 10,000 \\
ImageNet & 224x224 color & 1000 & 1,281,167 & 50,000 \\
\bottomrule
\end{tabular}
\end{center}
\end{table}

We performed 3 runs for each dataset, each run with a set of 5 different models.
Ensembles were generated with \textit{topk} values of $[1, 2, 3]$.
Results are shown in Table~\ref{tab:dl-results}.

\begin{table}[hbt!]
\begin{center}
\caption{Accuracy results for DL runs: test-set accuracy of best single model vs. Classy Ensemble (the winner is boldfaced).
Models: set of 5 models.
Best: best single model test-set accuracy, with model in parentheses.
Classy: best Classy Ensemble test-set accuracy, with best \textit{topk} in parentheses.\\
\texttt{\scriptsize
Pretrained1: \{efficientnet\_b3, efficientnet\_b4, efficientnet\_b5, efficientnet\_b6, efficientnet\_b7\};\\
Pretrained2: \{vgg13\_bn, vgg16, vgg16\_bn, vgg19, vgg19\_bn\};\\
Pretrained3: \{resnet18, resnet34, resnet50, resnet101, resnet152\}
}
}
\label{tab:dl-results}%
\begin{tabular}{@{}r@{\hskip 10pt}c@{\hskip 10pt}c@{\hskip 13pt}c@{\hskip 13pt}c@{}}
\toprule
Dataset & Models & Best & Classy \\ 
\midrule
\multirow{3}{*}{Fashion MNIST} 
                & Pretrained1 & 92.4\% (efficientnet\_b6) & \textbf{93.0\%} (3) \\
                & Pretrained2 & 94.2\% (vgg16\_bn) & \textbf{94.7\%} (2) \\
                & Pretrained3 & 92.3\% (resnet34) & \textbf{93.6\%} (3) \\ \hline 

\multirow{3}{*}{CIFAR10}
                & Pretrained1 & 77.0\% (efficientnet\_b7) & \textbf{79.9\%} (2) \\
                & Pretrained2 & 87.8\% (vgg16\_bn) & \textbf{89.8\%} (2) \\
                & Pretrained3 & 76.2\% (resnet18) & \textbf{79.4\%} (2) \\ \hline 

\multirow{3}{*}{CIFAR100}
                & Pretrained1 & 48.8\% (efficientnet\_b7) & \textbf{53.0\%} (3) \\
                & Pretrained2 & 61.7\% (vgg16\_bn) & \textbf{68.4\%} (3) \\
                & Pretrained3 & 48.7\% (resnet34) & \textbf{52.5\%} (2) \\ \hline 

\multirow{3}{*}{ImageNet}
                & Pretrained1 & 79.3\% (efficientnet\_b4) & \textbf{80.3\%} (2) \\
                & Pretrained2 & \textbf{74.2\%} (vgg19\_bn) & 74.0\% (3) \\
                & Pretrained3 & 78.3\% (resnet152) & \textbf{78.9\%} (3) \\
\bottomrule
\end{tabular}
\end{center}
\end{table}

\textbf{Classy Evolutionary Ensemble.}
We observed in Table~\ref{tab:dl-results} that different sets of models given to Classy Ensemble to ``pick from'' yielded quite different results. We asked whether a good set could be derived algorithmically, instead of pre-specified manually (Pretrained1/2/3 in Table~\ref{tab:dl-results}). Towards this end we devised a simple evolutionary algorithm, using the recently introduced EC-KitY package \cite{eckity2023}. 

We focused on ImageNet, somewhat of a ``holy grail'' of DL classification, and undoubtedly significantly more difficult than the other 3 datasets in Table~\ref{tab:dl-results}. We also went beyond the PyTorch pretrained models, obtaining better-performing models through \texttt{timm}---PyTorch Image Models \cite{rw2019timm}.

The population comprised binary vectors, each representing a set of models given to Classy Ensemble, with a value of $1$ in position $i$ denoting inclusion of model $i$, and a value of $0$ denoting exclusion of model $i$ (of a list of 30 models given to the evolutionary algorithm as input). 

To compute the fitness of a single individual in the population, the set of models it represented was given to Classy Ensemble (Algorithm~\ref{alg:classy}) and the generated ensemble was tested on ImageNet's train set (we used a subset of the full train set to afford more runs).
We experimented with 3 fitness functions:
\begin{enumerate}
    \item $\mathit{accuracy}$,
    \item $\mathit{accuracy} + 1/ensemble\,size$ (to encourage small ensembles),
    \item $\mathit{accuracy} - \mathit{cosine\,similarity\,of\,ensemble\,output\,vectors}$ (to encourage diverse ensembles).
\end{enumerate}
Each evolutionary run used one of the 3 fitness functions, as well as a pre-set value of \textit{topk} $\in [1,2,3,4,5,6,7]$. 

We performed a total of 63 evolutionary runs---3 per each $(\mathit{fitness},\,\mathit{topk}$) pair---with the (main) hyperparemeters being:
population size -- 200, 
generation count -- 20,
tournament selection with tournament size 3,
single-point crossover with probability 0.5,
and bitwise mutation with probability 0.05.

The best single-model test-set score was 86.1\%, attained by model \newline
\texttt{\small beitv2\_base\_patch16\_224}. 
Classy Evolutionary Ensemble improved this to 86.4\% (with \textit{topk}$=5$ and the first fitness function) by finding the following set of models:
\newline
\texttt{\small
\{maxvit\_xlarge\_224, beitv2\_base\_patch16\_224, xcit\_large\_24\_p16\_224\_dist,\\
swinv2\_cr\_large\_224, vit\_huge\_patch14\_224, deit3\_huge\_patch14\_224\}}.

\section{Concluding Remarks}
\label{sec:conc}

We presented a novel ensemble-generation algorithm that takes a collection of fitted models and creates a potent ensemble. We demonstrated the efficacy of our algorithm by running experiments over 153 datasets.

We note that the implementation of Classy Ensemble is fairly simple (and, as noted above, the code is available at \url{https://github.com/...}). It is straightforward to incorporate our method in one's setup, using models that are often available anyway, as part of multiple runs.

We then enhanced our method to work with state-of-the-art DL models, showing that improvements could be attained.

Finally, when learning was combined with evolution, results improved on the hardest dataset.

In the future we plan to continue work on improving the algorithm, e.g., use class-dependent weighting, which may prove useful where imbalanced datasets are concerned. We could perform a more in-depth analysis of datasets for which Classy Ensemble is successful. We also wish to further explore the merit of Classy Evolutionary Ensemble in deep network settings.

\section*{Acknowledgments}
I thank Raz Lapid for helpful comments.

\bibliographystyle{splncs04}
\bibliography{bibliography}

\end{document}